\documentclass{bmcart}

\usepackage{amsthm,amsmath}
\usepackage[utf8]{inputenc} %unicode support
\usepackage{amssymb} % for itemsymbol - vartriangleright
\usepackage{algorithm}
\usepackage{algorithmic}
\usepackage{amssymb} % for itemsymbol - vartriangleright
\usepackage{amsmath} % for subequations
\usepackage{wrapfig}
\usepackage{graphicx}
\usepackage{multirow}
\usepackage{colortbl,booktabs} % color table rows
\usepackage{lipsum}   % dummy text
\usepackage{amsthm} % for proof environment
\usepackage{amsmath} % for split environment
\usepackage{braket}
\usepackage{hyperref}
\usepackage{cite}
\usepackage{adjustbox}
\usepackage{float} % for forcing figure placement

\newtheorem{theorem}{Theorem}[section]

\newtheorem{assumption}{Assumption}[section]

\newtheorem{remark}{Remark}[section]

\usepackage{graphicx}

\def\ROWCOLOR{black!18!white}

\DeclareMathOperator*{\argmin}{argmin}

\newcommand{\limk}{\lim_{k\rightarrow\infty}}
\newcommand{\prox}[1]{\mathrm{prox}_{#1} }

\newcommand{\fix}[1]{\mathrm{fix}(#1)}

\newcommand{\sA}{{\cal A}} 
 
\newcommand{\sC}{{\cal C}} 
\newcommand{\sD}{{\cal D}}

\newcommand{\sJ}{{\cal J}}

\newcommand{\sN}{{\cal N}}

\newcommand{\sS}{{\cal S}} 
  
\newcommand{\sU}{{\cal U}}

\newcommand{\bbN}{{\mathbb N}} 
 
\newcommand{\bbR}{{\mathbb R}}
\newcommand{\bbE}{{\mathbb E}}
\newcommand{\bbP}{{\mathbb P}}

\newcommand{\ie}{{\textit{i.e.}\ }}
\newcommand{\eg}{{\textit{e.g.}\ }}

\usepackage{wrapfig}

\usepackage{tikz}
\usepackage{subfig}
\usetikzlibrary{tikzmark,calc,,arrows,shapes,decorations.pathreplacing}
\tikzset{every picture/.style={remember picture}}
\usetikzlibrary{fit,shapes.misc}
\usetikzlibrary{positioning,backgrounds,spy}

\usepackage{hyperref}
\definecolor{bl}{RGB}{30,30,150}
\hypersetup{colorlinks=true, citecolor=bl, linkcolor=bl, urlcolor=bl}
\startlocaldefs
\endlocaldefs

\begin{document}

%%% Start of article front matter
\begin{frontmatter}

\begin{fmbox}
\dochead{Research}

\title{Feasibility-based Fixed Point Networks}

%%%%%%%%%%%%%%%%%%%%%%%%%%%%%%%%%%%%%%%%%%%%%%
%%                                          %%
%% Enter the authors here                   %%
%%                                          %%
%% Specify information, if available,       %%
%% in the form:                             %%
%%   <key>={<id1>,<id2>}                    %%
%%   <key>=                                 %%
%% Comment or delete the keys which are     %%
%% not used. Repeat \author command as much %%
%% as required.                             %%
%%                                          %%
%%%%%%%%%%%%%%%%%%%%%%%%%%%%%%%%%%%%%%%%%%%%%%

% \author[
%   addressref={aff1},                   % id's of addresses, e.g. {aff1,aff2}
%   corref={aff1},                       % id of corresponding address, if any
% % noteref={n1},                        % id's of article notes, if any
%   email={jane.e.doe@cambridge.co.uk}   % email address
% ]{\inits{J.E.}\fnm{Jane E.} \snm{Doe}}
% \author[
%   addressref={aff1,aff2},
%   email={john.RS.Smith@cambridge.co.uk}
% ]{\inits{J.R.S.}\fnm{John R.S.} \snm{Smith}}

%-------------------------
%-------------------------
\author[
  addressref={aff1},                   % id's of addresses, e.g. {aff1,aff2}
  corref={aff1},                       % id of corresponding address, if any
% noteref={n1},                        % id's of article notes, if any
  email={hheaton@ucla.edu}   % email address
]{\inits{H.H.}\fnm{Howard} \snm{Heaton}}
\author[
  addressref={aff1},                   % id's of addresses, e.g. {aff1,aff2}
  corref={aff1},                       % id of corresponding address, if any
% noteref={n1},                        % id's of article notes, if any
  email={swufung@math.ucla.edu}   % email address
]{\inits{S.W.F}\fnm{Samy} \snm{Wu Fung}}
\author[
  addressref={aff2},                   % id's of addresses, e.g. {aff1,aff2}
  corref={aff2},                       % id of corresponding address, if any
% noteref={n1},                        % id's of article notes, if any
  email={avivg@braude.ac.il}   % email address
]{\inits{A.G.}\fnm{Aviv} \snm{Gibali}}
\author[
  addressref={aff1},                   % id's of addresses, e.g. {aff1,aff2}
%   corref={aff1},                       % id of corresponding address, if any
% noteref={n1},                        % id's of article notes, if any
  email={wotaoyin@math.ucla.edu}   % email address
]{\inits{W.Y}\fnm{Wotao} \snm{Yin}}

%%%%%%%%%%%%%%%%%%%%%%%%%%%%%%%%%%%%%%%%%%%%%%
%%                                          %%
%% Enter the authors' addresses here        %%
%%                                          %%
%% Repeat \address commands as much as      %%
%% required.                                %%
%%                                          %%
%%%%%%%%%%%%%%%%%%%%%%%%%%%%%%%%%%%%%%%%%%%%%%

\address[id=aff1]{%                           % unique id
  \orgdiv{Department of Mathematics},             % department, if any
  \orgname{University of California, Los Angeles},          % university, etc
  \city{Los Angeles},                              % city
  \cny{United States}                                    % country
}

\address[id=aff2]{%
  \orgdiv{Department of Mathematics},
  \orgname{ORT Braude College},
  %\street{},
  %\postcode{}
  \city{Karmiel},
  \cny{Israel}
}
% \address[id=aff2]{%
%   \orgdiv{Institute of Biology},
%   \orgname{National University of Sciences},
%   %\street{},
%   %\postcode{}
%   \city{Kiel},
%   \cny{Germany}
% }

%%%%%%%%%%%%%%%%%%%%%%%%%%%%%%%%%%%%%%%%%%%%%%
%%                                          %%
%% Enter short notes here                   %%
%%                                          %%
%% Short notes will be after addresses      %%
%% on first page.                           %%
%%                                          %%
%%%%%%%%%%%%%%%%%%%%%%%%%%%%%%%%%%%%%%%%%%%%%%

%%\note{Sample of title note}     % note to the article
%\note[id=n1]{Equal contributor} % note, connected to author

\end{fmbox}% comment this for two column layout

%%%%%%%%%%%%%%%%%%%%%%%%%%%%%%%%%%%%%%%%%%%%%%%
%%                                           %%
%% The Abstract begins here                  %%
%%                                           %%
%% Please refer to the Instructions for      %%
%% authors on https://www.biomedcentral.com/ %%
%% and include the section headings          %%
%% accordingly for your article type.        %%
%%                                           %%
%%%%%%%%%%%%%%%%%%%%%%%%%%%%%%%%%%%%%%%%%%%%%%%

\begin{abstractbox}
\begin{abstract}
Inverse problems consist of recovering a signal from a collection of noisy measurements. These problems can often be cast as feasibility problems; however, additional regularization is typically necessary to ensure accurate and stable recovery with respect to data perturbations. 
Hand-chosen analytic regularization can yield desirable theoretical guarantees, but such approaches have limited effectiveness recovering signals due to their inability to leverage large amounts of available data. To this end, this work fuses data-driven regularization and convex feasibility in a theoretically sound manner. This is accomplished using feasibility-based fixed point networks (F-FPNs). Each F-FPN defines a collection of  nonexpansive operators, each of which is the composition of a   projection-based operator and a data-driven regularization operator. Fixed point iteration is used to compute fixed points of these  operators, and weights of the operators are tuned so that the fixed points closely represent available data. Numerical examples demonstrate   performance increases by F-FPNs when compared to standard TV-based recovery methods for CT reconstruction and a comparable neural network based on algorithm unrolling.\footnotemark
\end{abstract}

%%%%%%%%%%%%%%%%%%%%%%%%%%%%%%%%%%%%%%%%%%%%%%
%%                                          %%
%% The keywords begin here                  %%
%%                                          %%
%% Put each keyword in separate \kwd{}.     %%
%%                                          %%
%%%%%%%%%%%%%%%%%%%%%%%%%%%%%%%%%%%%%%%%%%%%%%


% MSC classifications codes, if any
%\begin{keyword}[class=AMS]
%\kwd[Primary ]{}
%\kwd{}
%\kwd[; secondary ]{}
%\end{keyword}

\end{abstractbox}
%
%\end{fmbox}% uncomment this for two column layout

\end{frontmatter} 

%%%%%%%%%%%%%%%%%%%%%%%%%%%%%%%%%%%%%%%%%%%%%%%%
%%                                            %%
%% The Main Body begins here                  %%
%%                                            %%
%% Please refer to the instructions for       %%
%% authors on:                                %%
%% https://www.biomedcentral.com/getpublished %%
%% and include the section headings           %%
%% accordingly for your article type.         %%
%%                                            %%
%% See the Results and Discussion section     %%
%% for details on how to create sub-sections  %%
%%                                            %%
%% use \cite{...} to cite references          %%
%%  \cite{koon} and                           %%
%%  \cite{oreg,khar,zvai,xjon,schn,pond}      %%
%%                                            %%
%%%%%%%%%%%%%%%%%%%%%%%%%%%%%%%%%%%%%%%%%%%%%%%%

%%%%%%%%%%%%%%%%%%%%%%%%% start of article main body
% <put your article body there> 

%%%%%%%%%%%%%%%%
%% Background %%
%%
\section{Introduction}
    \footnotetext{{\footnotesize Codes are available on Github: \href{https://github.com/howardheaton/feasibility_fixed_point_networks}{github.com/howardheaton/feasibility\_fixed\_point\_networks}}}
   Inverse problems arise in numerous applications such as medical imaging~\cite{arridge1999optical,arridge2009optical,hansen2006deblurring,osher2005iterative}, phase retrieval~\cite{bauschke2002phase,candes2015phase,fung2020multigrid}, geophysics~\cite{bui2013computational, fung2019multiscale,fung2019uncertainty,haber2000fast,haber2004inversion,kan2020pnkh}, and machine learning~\cite{cucker2002best,fung2019large,haber2017stable,vito2005learning,fung2020admm}. The   goal of inverse problems is to recover a signal\footnote{\footnotesize While we refer to signals, this phrase is generally meant to describe objects of interest that can be represented mathematically (\eg \footnotesize  images, parameters of a differential equation, and points in a Euclidean space).} $u_d^\star$ from a collection of indirect noisy measurements $d$.
   These quantities are typically related by a linear mapping $A$ via 
    \begin{equation}
        d = A u_d^\star  + \varepsilon,        
    \end{equation}
    where  $\varepsilon $ is   measurement noise. Inverse problems are often ill-posed, making   recovery of the signal $u_d^\star$   unstable for noise-affected data $d$. To overcome this,   traditional approaches estimate the   signal $u_d^\star$ by a solution $\tilde{u}_d$ to the variational problem 
    \begin{equation}
        \min_{u  } \; \ell(A u, d) + J(u),
        \label{eq: variational-problem}
    \end{equation}
    where $\ell $ is a fidelity term that measures the discrepancy between the measurements and the application of the forward operator $A$ to the signal estimate (\eg least squares). The function $J$ serves as a regularizer, which ensures both that the solution to (\ref{eq: variational-problem}) is unique and that its computation is stable. 
    In addition to ensuring well-posedness, regularizers are constructed in an effort to instill prior knowledge of the true signal, \eg sparsity $J(u) = \|u\|_1$~\cite{beck2009fast,candes2006quantitative,candes2006robust,donoho2006compressed}, Tikhonov $J(u) = \|u\|^2$~\cite{calvetti2003tikhonov,golub1999tikhonov}, total variation (TV) $J(u) = \| \nabla u \|_1$~\cite{chan2020two,rudin1992nonlinear}, and, more recently, data-driven regularizers~\cite{adler2018learned,kobler2017variational,lunz2019adversarial}.   
    A further generalization of using data-driven regularization consists of Plug-and-Play (PnP) methods~\cite{chan2016plug,cohen2020regularization,venkatakrishnan2013Plug}, which replace the proximal operators in an optimization algorithm with previously trained denoisers. \\
      
    An underlying theme of regularization is that signals represented in high dimensional spaces often exhibit a common structure.   Although hand picked regularizers may admit desirable theoretical properties   leveraging \textit{a priori} knowledge, they are typically unable to leverage available data. An ideal regularizer will leverage available data to  best capture the core properties that should be exhibited by output reconstruction estimates  of  true signals. Neural networks have demonstrated great success in this regard, achieving state of the art results~\cite{xu2014deep,jin2017deep}. However, purely data-driven machine learning approaches do little to leverage the underlying physics of a problem, which can lead to poor compliance with data \cite{moeller2019controlling}.
    On the other hand, fast  feasibility-seeking algorithms (\eg see \cite{censor2012effectiveness,censor2008diagonally,gordon2005component,censor2008iterative,censor2015projection} and references therein) efficiently leverage known physics   to solve   inverse problems, being able to handle massive-scale sets of constraints \cite{bauschke2013projection,censor2012effectiveness,ordonez2017realtime,penfold2010block}.   
    Thus, a relatively untackled question remains: \newline
    \begin{center}
        \textit{Is it possible to fuse   feasibility-seeking algorithms with data-driven regularization in a manner that improves reconstructions and yields convergence?}    \newline     
    \end{center} 
    This work answers the above inquiry affirmatively.
    The key idea is to use machine learning techniques to create a mapping $T_\Theta$, parameterized by weights $\Theta$. For fixed measurement data $d$, $T_\Theta(\cdot\ ;d)$ forms an operator possessing standard properties used in feasibility algorithms.  
    Fixed point iteration is used to find fixed points of $T_\Theta(\cdot\ ;d)$ and the weights $\Theta$ are tuned such that these fixed points both resemble available signal data and are consistent with measurements (up to the noise level).

    \paragraph{Contribution}  The core contribution of this work is to  connect  powerful feasibility-seeking algorithms to data-driven regularization in a manner that maintains theoretical guarantees. This is accomplished by presenting a feasibility-based fixed point network (F-FPN) framework that solves a learned feasibility problem. Numerical examples are   provided that demonstrate notable performance benefits to our proposed formulation when compared to TV-based methods and fixed-depth neural networks formed by algorithm unrolling.

\paragraph{Outline} We first overview convex feasibility problems (CFPs) and  a learned feasibility problem (Section \ref{sec: problem-formulation}).  Relevant neural network material is discussed next (Section \ref{sec: FPN}), followed by our proposed F-FPN framework   (Section  \ref{sec: proposed-method}). Numerical examples are  then provided   with discussion and conclusions (Sections \ref{sec: experiments} and \ref{sec: conclusion}).

\section{Convex Feasibility Background} \label{sec: problem-formulation}
\subsection{Feasibility Problem} 
Convex feasibility problems (CFPs) arise   in many real-world applications, \eg  imaging, sensor networks, radiation therapy treatment planning (see \cite{censor2012effectiveness, bauschke2015projection, bauschke2017convex} and the references therein).
We formalize the CFP setting and relevant methods as follows.
Let $\sU$ and $\sD$ be finite dimensional Hilbert spaces,   referred to as the signal and data spaces, respectively.\footnote{{\footnotesize The product and norm are denoted by $\left<\cdot,\cdot\right>$ and $\|\cdot\|$ respectively. Although we use the same notation for each space, it will be clear from context which one is   used.}} 
Given additional knowledge about a linear inverse problem, measurement data $d\in \sD$ can   be used to express a CFP solved by the true signal $u_d^\star\in\sU$   when measurements are noise-free. That is,       data $d$ can be used to define a collection $\{\sC_{d,\ell}\}_{\ell=1}^m$ of closed convex subsets of $\sD$ (\eg hyperplanes) such that the true signal $u_d^\star$ is contained in their intersection, \ie $u_d^\star$ solves the   problem\vspace{-5pt} 
\begin{equation}
    \mbox{Find $u_d$ such that\  } u_d \in \sC_d \triangleq \bigcap_{\ell=1}^m \sC_{d,\ell}.
    \tag{CFP}
    \label{eq: cfp}
\end{equation}
A common approach to solving (\ref{eq: cfp}), \textit{inter alia},   is to use projection algorithms \cite{bauschke1996projection}, which utilize orthogonal projections onto the individual sets $\sC_{d,\ell}$. For  a closed, convex, and nonempty set $\sC\subseteq \sU$, the   projection $P_\sC:\sU\rightarrow \sC$ onto   $\sC$,  is  defined by\\[-15pt]
\begin{equation}
    P_\sC(u) \triangleq \argmin_{v\in \sC} \dfrac{1}{2}\|v-u\|^2.
    \label{eq: projection-def}
\end{equation}
Projection algorithms are iterative in nature and each update uses combinations of projections onto each set $\sC_{d,\ell}$, relying on the principle that it is generally much easier to project onto the individual sets than onto their intersection.  
%Such modelling reformulation enables to use, {\it inter alia}, the so-called projection methods, which have enjoyed great computational success in various applied problems  \cite{bauschke2015projection,bauschke2017convex}. 
These methods date back to the 1930s  \cite{kaczmarz1937angenaherte,cimmino1938cacolo} and have been adapted to now handle huge-size problems of dimensions for which more sophisticated methods cease to be efficient or even applicable due to memory requirements \cite{censor2012effectiveness}. Computational simplicity derives from the fact the building bricks of a projection algorithm are the projections onto   individual sets. Memory efficiency occurs because the algorithmic structure is either sequential or simultaneous (or hybrid) as in the block-iterative projection methods \cite{aharoni1989block,byrne1996block}  and string-averaging projection methods  \cite{censor2012effectiveness,censor2013convergence,censor2009string,censor2003convergence}.  
These algorithms   generate sequences that solve (\ref{eq: cfp}) asymptotically, and the update operations can be iteration dependent (\eg cyclic   projections). 
We let $\sA_d^k$ be the update operator for the $k$-th step of a projection algorithm  solving (\ref{eq: cfp}). Consequently, each projection algorithm generates a sequence $\{u^k\}_{k\in\bbN}$  via the fixed point iteration
\begin{equation}
    u^{k+1} \triangleq \sA_d^k(u^k), \ \ \ \mbox{for all $k\in\bbN$.}
    \tag{FPI}
    \label{eq: fixed-point-iteration}
\end{equation}
A common assumption for such methods is   the intersection of all the algorithmic operators' fixed point sets\footnote{{\footnotesize For an operator $T$, its fixed point set is $\fix{T}\triangleq \{u: u=T(u)\}$.}} contains or forms the desired set $\sC_d$, \ie
\begin{equation}
    \sC_d = \bigcap_{k=1}^\infty \fix{\sA_d^k},
\end{equation} 
which automatically holds when  $\{\sA_d^k\}_{k\in\bbN}$ cycles over a collection of projections. 

\subsection{Data-Driven Feasibility Problem}
As noted previously, inverse problems are often ill-posed, making    (\ref{eq: cfp}) insufficient to faithfully recover the signal $u_d^\star$. Additionally, when noise is present, it can often be the case that the intersection is empty (\ie $\sC_d = \emptyset$). This calls for a different model to recover $u_d^\star$.
To date, projection methods have limited inclusion of regularization (\eg superiorization~\cite{davidi2009perturbation,censor2010perturbation,herman2012superiorization,he2017perturbation,censor2014weak}, sparsified Kaczmarz \cite{schopfer2019linear,lorenz2014sparse}). Beyond sparsity via $\ell_1$ minimization, such approaches typically do not yield   guarantees beyond feasibility (\eg it may be desirable to minimize a regularizer over $\sC_d$). 
We propose composing a projection algorithm and a data-driven regularization operator in a manner so   each update is analogous to a proximal-gradient step.
This is accomplished via   a parameterized mapping $R_\Theta:\sU\rightarrow\sU$, with weights\footnote{\footnotesize Operator weights are also commonly called parameters.} denoted by $\Theta$. This   mapping directly leverages available data (explained in Section \ref{sec: FPN}) to   learn features shared among   true signals of interest.   
We   augment (\ref{eq: cfp}) by   using   operators $\{\sA_k\}_{k\in\bbN}$ for solving (\ref{eq: cfp}) and instead solve the learned common fixed points (\ref{eq: L-CFP}) problem 
\begin{equation}
    \mbox{Find $\tilde{u}_d$ such that\ } \tilde{u}_d \in \sC_{\Theta,d} \triangleq  \bigcap_{k=1}^\infty \fix{\sA_d^k \circ R_\Theta}.
    \label{eq: L-CFP}
    \tag{L-CFP}
\end{equation} 
Loosely speaking, when $R_\Theta$ is chosen  well,  the signal $\tilde{u}_{d}$ closely approximates $u_d^\star$. \\

We utilize classic operator results to solve (\ref{eq: L-CFP}). 
An operator $T\colon \sU\rightarrow \sU $ is \textit{nonexpansive} if it is 1-Lipschitz, \ie
\begin{equation}
    \|T(u) - T(u)\|\leq \|u-v\|, \ \ \ \mbox{for all $u,v\in D$.}
\end{equation}
Also, $T$ is \textit{averaged} if there exists $\alpha \in (0,1)$ and a nonexpansive operator $Q:\sU\rightarrow\sU$ such that
$
    T(u) = (1-\alpha) u + \alpha Q(u)
$ for all $u\in \sU$.
For example, the projection $P_\sS$ defined in (\ref{eq: projection-def}) is averaged along with convex combinations of projections \cite{cegielski2012iterative}.
Our method utilizes the following standard assumptions, which are typically satisfied by projection methods (in the noise-free setting with $R_\Theta$ as the identity).

\begin{assumption} \label{ass: NE-nonempty}
    The intersection set $\sC_{\Theta,d}$ defined in (\ref{eq: L-CFP}) is nonempty and $\{(\sA_d^k\circ R_\Theta)\}_{k\in\bbN}$ forms a sequence of nonexpansive operators.
\end{assumption}

\begin{assumption} \label{ass: asymptotically-regular}
    For any sequence $\{u^k\}_{k\in\bbN}\subset \sU$, the sequence of operators $\{(\sA_d^k\circ R_\Theta)\}_{k\in\bbN}$  has the property
    \begin{equation}
        \limk \|(\sA_d^k\circ R_\Theta)(u^k)-u^k\| = 0
        \ \ \ \Longrightarrow \ \ \ 
        \liminf_{k\rightarrow\infty} \|P_{\sC_{\Theta,d}}(u^k)-u^k\| = 0.
    \end{equation} 
\end{assumption}

When a finite collection of update operations are used and applied (essentially) cyclically, the previous assumption automatically holds (\eg  setting  $\sA_d^k \triangleq P_{\sC_{d,i_k}}$ and $i_k \triangleq k \ \mbox{mod}(m) + 1$). 
We use the learned fixed point iteration to solve (\ref{eq: L-CFP})  
\begin{equation}
    u^{k+1} \triangleq ( \sA_d^k\circ R_\Theta)(u^k), \ \ \ \mbox{for all $k\in\bbN$.}
    \tag{L-FPI}
    \label{eq: L-FPI}
\end{equation}
Justification of the (\ref{eq: L-FPI}) iteration is provided by the following theorems, which are rewritten from their original form to a manner that matches the present context.

\begin{theorem} 
\label{thm: KM}
{\sc (Krasnosel'ski\u{\i}-Mann \cite{Krasnoselskii1955_two,Mann1953_mean})}
If $(\sA_d\circ R_\Theta)\colon \sU\rightarrow\sU$ is averaged and contains a fixed point, then, for any $u^1\in\sU$, the sequence $\{u^k\}_{k\in\bbN}$ generated by (\ref{eq: L-FPI}), taking $\sA_d^k \circ R_\Theta = \sA_d\circ R_\Theta$, converges to a fixed point of $\sA_d\circ R_\Theta$.
\end{theorem} 

\begin{theorem} {\sc  (Cegieslki \cite{cegielski2012iterative})}
    If Assumptions \ref{ass: NE-nonempty} and \ref{ass: asymptotically-regular} hold, and if $\{u^k\}_{k\in\bbN}$ is a sequence generated by the iteration  (\ref{eq: L-FPI}) satisfying $\|u^{k+1}-u^k\|\rightarrow 0$, then $\{u^k\}_{k\in\bbN}$  converges to a limit $u^\infty \in \sC_{\Theta,d}$.
\end{theorem}
 
 \begin{figure}[t] 
    \centering
    \includegraphics[width=0.95\textwidth]{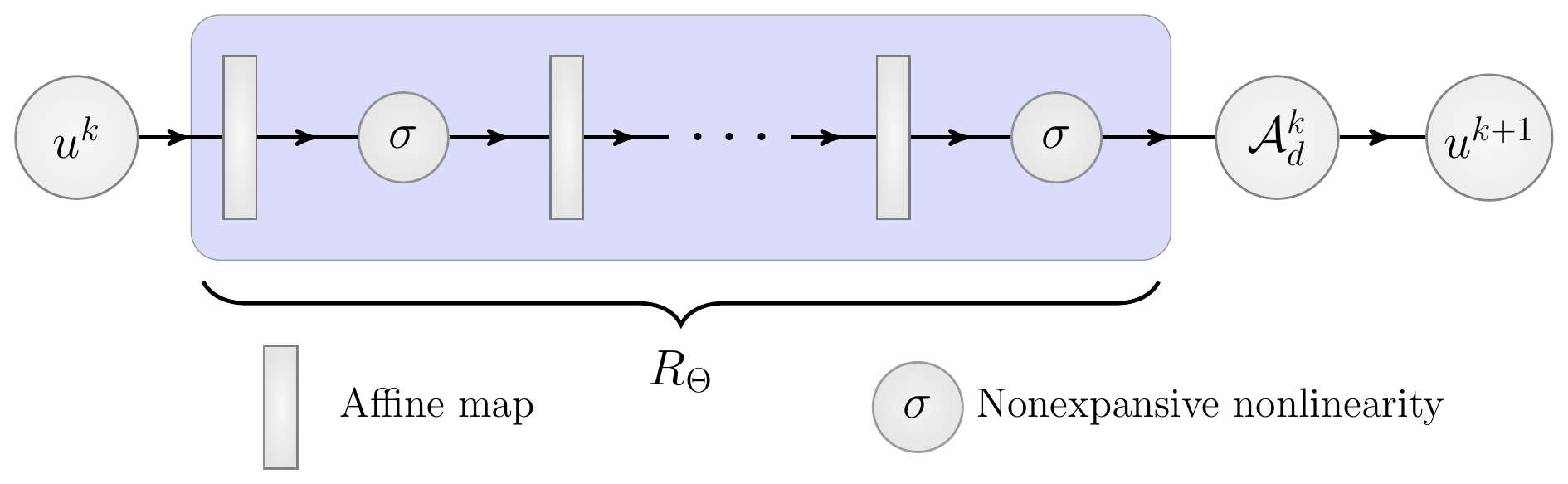}
    \caption{Diagram for update operations in the learned fixed point iteration (\ref{eq: L-FPI}) to solve (\ref{eq: L-CFP}). Here $R_\Theta$ is comprised of a finite sequence of applications of (possibly) distinct affine mappings (\eg convolutions), and nonlinearities (\eg projections on the nonnegative orthant, \ie ReLUs).
    For each $k\in\bbN$, we let $\sA_d^k$ be a  projection-based algorithmic operator. The  parameters $\Theta$ of $R_\Theta$ are tuned in an offline process by solving (\ref{eq: training-problem}) to ensure signals are faithfully recovered.  }
    \label{fig: T-diagram}
\end{figure}
\vspace{5pt}
 
\section{Fixed Point Networks Overview} \label{sec: FPN}
One of the most promising areas in artificial intelligence is deep learning,
a form of machine learning that uses neural networks containing many hidden layers~\cite{lecun2015deep,bengio2009learning}. 
Deep learning tasks in the context of this work can be cast as follows. 
Given   measurements $d$ drawn from a distribution  $\bbP_\sD$ and  corresponding  signals $u_d^\star$ drawn from a distribution $\bbP_\sU$, we seek a mapping $\sN_\Theta \colon \sD  \rightarrow \sU$ that approximates a one-to-one correspondence between the measurements and signals, \ie  
\begin{equation}
    \label{eq: f_equals_y} 
    \sN_\Theta(d) \approx u_d^\star,\ \ \ \mbox{for all $d\sim \bbP_\sD$.}
\end{equation}
Depending on the nature of the given data, the task at hand can be regression or classification. 
In this work, we focus on solving regression problems where the learning is \emph{supervised}, \ie the loss function explicitly uses a correspondence between   input and output data pairings. 
When the loss function does not use this correspondence (or when not all data pairings are available),   the learning is \emph{semi-supervised} if partial pairings are used and \emph{unsupervised} if no pairings are used. 

\newpage
\subsection{Recurrent Neural Networks}
\label{subsec: RNNs}
% The mapping $\sN_\Theta$ can be thought of as an interpolation or approximation function.
A common model for $\sN_\Theta$ is given by Recurrent Neural Networks (RNNs)~\cite{rumelhart1986learning}, which have enjoyed a great deal of success in natural language processing (NLPs)~\cite{manning1999foundations}, time series~\cite{hastie2009elements}, and classification~\cite{hastie2009elements}. 
An $N$-layer RNN  takes observed data $d$ as input and can be modeled as the $N$-fold composition of a mapping $T_\Theta$ via
\begin{equation}
    \label{eq: RNN-def}
    \begin{split}
        \sN_\Theta \triangleq \underbrace{T_{\Theta} \circ T_{\Theta} \circ \ldots \circ T_{\Theta}}_{N \text{ times}}.
    \end{split}
\end{equation} 
Here $T_{\Theta}(u; d)$ is an operator comprised of a finite sequence of applications of (possibly) distinct affine mappings and nonlinearities, and $u$ is initialized to some fixed value (\eg the zero vector). 
To identify a faithful mapping $\sN_\Theta$ as in \eqref{eq: f_equals_y}, we solve a training problem. This is modeled as finding weights that minimize an expected loss, which is typically solved using optimization methods like  SGD~\cite{BottouCurtisNocedal2016_optimization} and Adam \cite{kingma2015adam}. In particular, we solve  the training problem
\begin{equation}
    \min_\Theta \;  \bbE_{d \sim \mathcal{D}} \left[ \ell(\sN_\Theta(d), u_d^\star)\right],
    \label{eq: training-problem}
\end{equation}
where $\ell \colon  \sU \times \sU \to \bbR$ models the discrepancy between the prediction $\sN_\Theta(d)$ of the network and the training data $u_d^\star$.
In practice, the expectation in (\ref{eq: training-problem}) is approximated using a finite subset of data, which is referred to as \textit{training data}.  In addition to minimizing over the training data, we aim for~\eqref{eq: f_equals_y} to hold for   a set of \emph{testing data} that was not used during training (which tests the network's ability to \emph{generalize}).  

\begin{remark}
 We emphasize a time intensive offline process is used to find a solution $\Theta^\star$ to  (\ref{eq: training-problem}) (as is common in machine learning). After this, in the online setting, we apply $\sN_{\Theta^\star}(d)$ to   recover a signal $u_d^\star$ from its previously unseen measurements $d$, which is a much faster process.  
\end{remark}

\begin{remark}
    If we impose a particular structure to $T_\Theta$, as shown in Figure~\ref{fig: T-diagram}, an $N$-layer RNN can be interpreted as an unrolled fixed point (or optimization) algorithm that runs for $N$ iterations. Our experiments compare our proposed method to such an unrolled scheme.
\end{remark}

\subsection{Fixed-Point Networks}
Increasing neural network depth leads to more expressibility \cite{fan2020universal,tabuada2020universal}.
A recent trend in deep learning seeks to inquires:  what happens when the number of recurrent layers $N$ goes to infinity? 
Due to ever growing memory requirements (growing linearly with $N$) to train networks, directly unrolling a sequence generated by successively applying $T_\Theta$ is, in general, intractable  for arbitrarily large $N$.
However, the sequence limit can be modeled using a fixed point equation. In this case, evaluating a fixed point network (FPN) \cite{fung2021fixed} is equivalent to finding the unique fixed point of an averaged operator $T_\Theta(\cdot\ ;d)$, \ie an FPN $\sN_\Theta$ is defined by\footnote{\footnotesize The presented definition is a slight variation of the original work, adapted to this setting.}
\begin{equation}
    \sN_\Theta(d) \triangleq u_{\Theta,d},
    \ \ \ \mbox{where} \ \ \ u_{\Theta,d} = T_\Theta(u_{\Theta,d};d).
    \label{eq: FPN}
\end{equation}
Standard results  \cite{browder1965nonexpansive,gohde1965prinzip,kirk1965fixed} can be used to guarantee existence\footnote{\footnotesize The original FPN paper used a more restrictive contraction condition to guarantee uniqueness and justify how the weights are updated during training. However, we use their method in our more general setting since the contraction factor can be arbitrarily close to unity.} of fixed points of nonexpansive $T_\Theta$.
Iteratively applying $T_\Theta$ produces a convergent sequence (Theorem \ref{thm: KM}). However, for different $d$, this procedure the number of steps to converge may vary, and so these models belong to the class of \emph{implicit depth models}. 
As mentioned, it is computationally intractable to differentiate $\ell$ with respect to $\Theta$ by applying the chain rule through each of  the $N$ layers (when $N$ is sufficiently large). Instead, the gradient $\mathrm{d}\ell/\mathrm{d}\Theta$   is computed via the implicit function theorem~\cite{krantz2012implicit}. 
Specifically, the gradient is obtained by solving the Jacobian-inverse equation (\eg see~\cite{bai2019deep,winston2020monotone,chen2018neural})
\begin{equation}
    \label{eq: implicit_func_thm}
    \frac{\mathrm{d}\ell}{\mathrm{d}\Theta} = \sJ_{\Theta}^{-1} \frac{\partial T}{\partial \Theta}, \quad \text{ where } \quad \sJ_\Theta \triangleq I - \dfrac{dT_\Theta}{d u}.
\end{equation}
% where $J_\Theta = I - \dfrac{dT_\Theta}{d u}$. 
Recent works that solve the Jacobian-inverse equation in~\eqref{eq: implicit_func_thm} to train neural networks include Deep Equilibrium Networks~\cite{bai2019deep,bai2020multiscale} and Monotone Equilibrium Networks~\cite{winston2020monotone}.
A key difficulty arises when computing the gradient via (\ref{eq: implicit_func_thm}), especially when the signal space has large dimensions (\eg  when $u_d^\star$ is a high-resolution image). Namely, a linear system involving the Jacobian term $\sJ_\Theta$ must be approximately solved to estimate the gradient of $\ell$.
Recently, a new framework for training implicit depth models, called Jacobian-Free Backpropagation (JFB)~\cite{fung2021fixed}, was presented in the context of FPNs, which avoids the intensive linear system solves at each step.
The idea is to replace gradient $\mathrm{d}\ell/\mathrm{d}\Theta$ updates with $\partial T/\partial\Theta$, which is equivalent to a preconditioned gradient (since $\sJ_\Theta^{-1}$ is coercive \cite[Lemma A.1]{fung2021fixed}). 
JFB  provides a descent direction and was found to be effective and competitive for training implicit-depth neural networks at substantially reduced computational costs. 
Since the present work solves inverse problems where the signal space has   very high dimension, we   leverage FPNs and JFB to solve~\eqref{eq: training-problem} for our proposed method.

\subsection{Learning to Optimize} An emerging field in machine learning is known as ``learning to optimize'' (L2O) (\eg see the survey works \cite{monga2021algorithm,chen2021learning}).
As a paradigm shift away from conventional optimization algorithm design, L2O uses machine learning to improve an optimization method.  
Two approaches are typically used for model-based algorithms.
Plug-and-Play (PnP) methods learn a denoiser in the form of a neural network and then plug this denoiser into an optimization algorithm (\eg to replace a proximal for total variation). Here training of the denoiser is separate from the task at hand.
On the other hand, unrolling methods incorporate tunable weights into an algorithm that is truncated to a fixed number of iterations, forming a neural network.
Unrolling the iterative soft thresholding algorithm (ISTA), the authors in \cite{gregor2010learning} obtained the first major L2O scheme Learned ISTA (LISTA) by letting each matrix in the updates be tunable.
Follow-up papers also demonstrate empirical success in various applications, include compressive sensing \cite{rick2017one,metzler2017learned,Chen_Liu_Wang_Yin_2018,Diamond_Sitzmann_Heide_Wetzstein_2018,Perdios_Besson_Rossinelli_Thiran_2017,mardani2018neural,zhang2018ista,Ito_Takabe_Wadayama_2019,mardani2019degrees}, denoising \cite{Diamond_Sitzmann_Heide_Wetzstein_2018,Putzky_Welling_2017,zhang2017learning,Chen_Pock_2017,sreter2018learned,LiuChenWangYin2019_alista,xie2019differentiable,lunz2018adversarial,mardani2019degrees}, and
deblurring \cite{Diamond_Sitzmann_Heide_Wetzstein_2018,zhang2017learning,meinhardt2017learning,liu2018bridging,Corbineau_Bertocchi_Chouzenoux_Prato_Pesquet_2019,mardani2019degrees,mukherjee2020learned,zhang2019deep,li2020efficient}.  
L2O schemes are related to our   method, but no L2O scheme has, to our knowledge, used a fixed point model as in (\ref{eq: L-CFP}). Additionally, our JFB training regime  differs from the L2O unrolling and PnP schemes.  \\

\newpage
 
\section{Proposed Method} \label{sec: proposed-method}
Herein we present the feasibility-based FPN (F-FPN).
Although based on FPNs, here we replace the single operator of FPNs  by a sequence of operators, each taking the form of a composition. Namely, we use updates in the iteration (\ref{eq: L-FPI}). 
The assumptions necessary for convergence can be approximately ensured  (\eg see Subsection \ref{subsec: approximate-contraction} in the Appendix). 
This iteration yields the F-FPN $\sN_\Theta$, defined by
\begin{equation}
    \sN_\Theta(d) \triangleq \tilde{u}_d,
    \ \ \ \mbox{where} \ \ \ 
    \tilde{u}_d = \bigcap_{k=1}^\infty \fix{\sA_d^k \circ R_\Theta},
\end{equation}
assuming the intersection is unique.\footnote{\footnotesize Uniqueness is unlikely in practice; however, this assumption is justified since we use the same initial iterate $u^1$ for each initialization. This makes recovery of the same signal is stable with respect to changes in $\Theta$.}
This is approximately implemented via Algorithm \ref{alg: F-FPN}.
     \begin{algorithm}[t]
        \caption{Feasiblity-based Fixed Point Network (F-FPN)}
        \label{alg: F-FPN}
        \begin{algorithmic}[1]           
            \STATE{\begin{tabular}{p{0.35\textwidth}r}
             \hspace*{-8pt} {\sc $\sN_{\Theta}(d)$:}
             & 
             $\vartriangleleft$ Input data is $d$
             \end{tabular}}    
            
            \STATE{\begin{tabular}{p{0.35\textwidth}r}
             \hspace*{2pt} $u^1\leftarrow \tilde{u}$
             & 
             $\vartriangleleft$ Initialize iterate to fixed reference
             \end{tabular}}

            \STATE{\begin{tabular}{p{0.35\textwidth}r}
             \hspace*{2pt} $k \leftarrow 1$
             & 
             $\vartriangleleft$ Initialize iteration counter
             \end{tabular}}                 

            \STATE{\begin{tabular}{p{0.35\textwidth}r}
             \hspace*{2pt} {\bf while $\|u^{k+1}-u^k\|\geq  \delta$}  
             & 
             $\vartriangleleft$ Loop to convergence
             \end{tabular}}  
             
            \STATE{\begin{tabular}{p{0.35\textwidth}r}
             \hspace*{12pt} $u^{k+1} \leftarrow (\sA_d^k \circ R_\Theta)(u^k;d)$
             & 
             $\vartriangleleft$ Apply regularization $R_\Theta$ and feasibility step $\sA_d^k$
             \end{tabular}}        
             
            \STATE{\begin{tabular}{p{0.35\textwidth}r}
             \hspace*{12pt} $k \leftarrow k + 1$
             & 
             $\vartriangleleft$ Increment counter
             \end{tabular}}                 
        
            \STATE{\begin{tabular}{p{0.35\textwidth}r}
             \hspace*{2pt} {\bf return} $u^k$
             & 
             $\vartriangleleft$ Output solution estimate 
             \end{tabular}}   
        \end{algorithmic}
\end{algorithm}  
The weights $\Theta$ of the network $\sN_\Theta$ are tuned by solving the   training problem (\ref{eq: training-problem}).  In an ideal situation, the optimal weights $\Theta^\star$ solving (\ref{eq: training-problem}) would yield feasible outputs (\ie $\sN_\Theta(d) \in \sC_d$ for all data $d\in\sC$) that also resemble the true signals $u_d^\star$. However, measurement noise in practice makes it unlikely that $\sN_\Theta(d)$ is feasible, let alone that $\sC_d$ is nonempty. 
In the noisy setting, this is no longer a concern since we augment (\ref{eq: cfp}) via (\ref{eq: L-CFP}) and are ultimately concerned with recovering a signal $u_d^\star$, not solving a feasibility problem. In summary, our model is based on the underlying physics of a problem (via the convex feasibility structure), but is also steered by available data via the training problem (\ref{eq: training-problem}). Illustrations of the efficacy of this approach are provided in Section \ref{sec: experiments}.

\begin{figure}[t]
      \centering
      \small
        \setlength{\tabcolsep}{7pt}
        \begin{tabular}{c c c}    
        
        \includegraphics[width=0.25\textwidth]{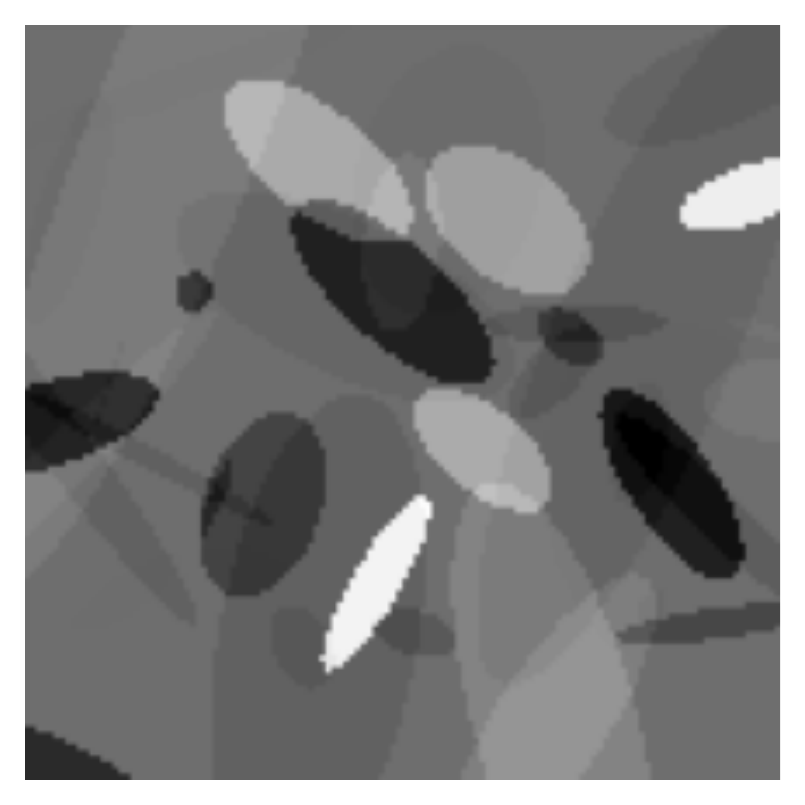}       
        &  
        \includegraphics[width=0.25\textwidth]{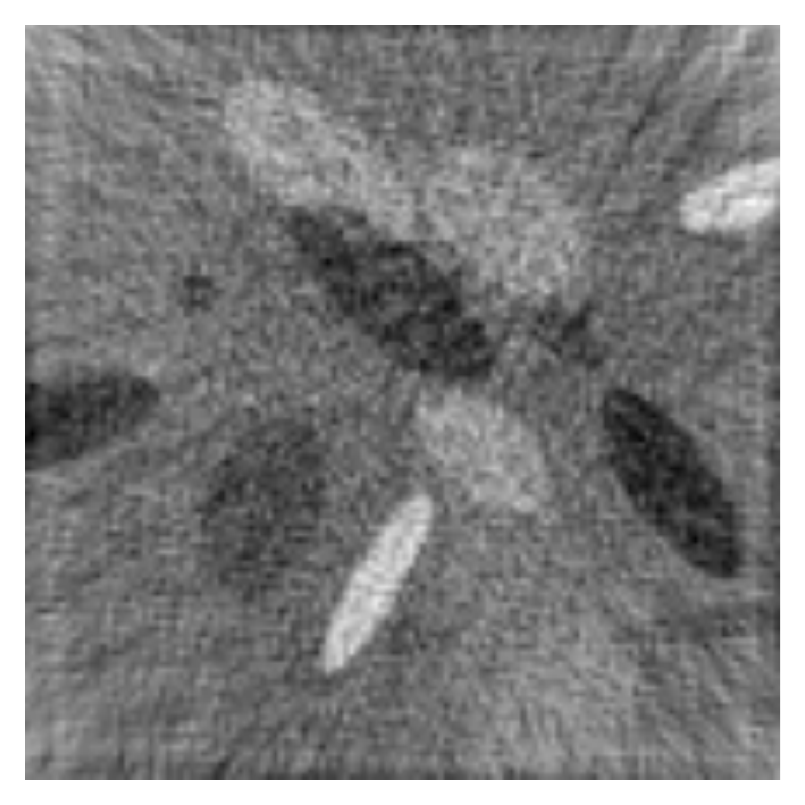}      
        & 
        \includegraphics[width=0.25\textwidth]{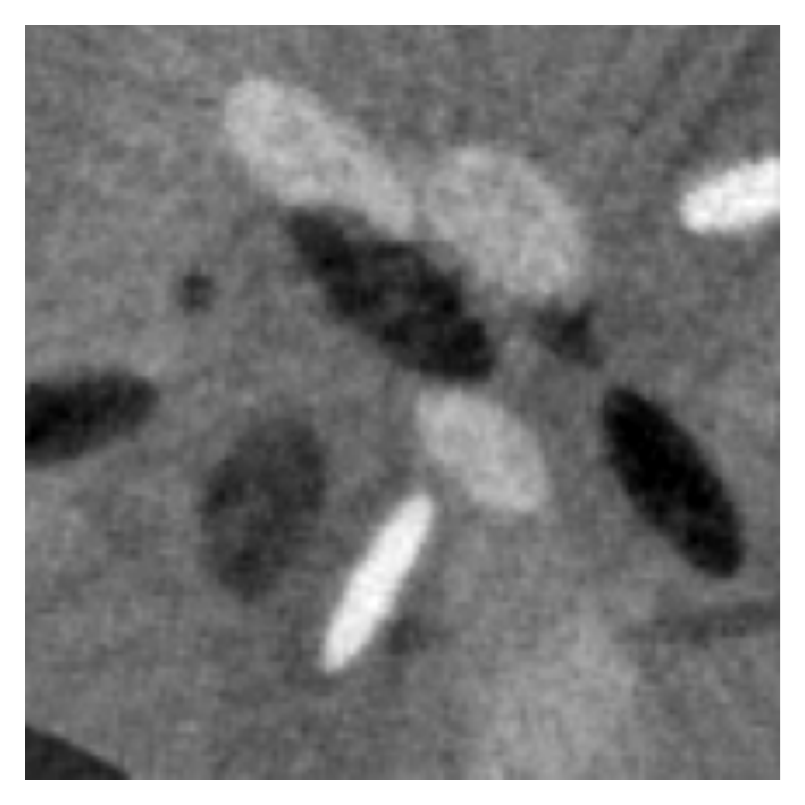}    \\
        
            {\bf Ground Truth} & {\bf FBP} &{\bf  TVS}
            \\      
            SSIM: 1.000 & SSIM: 0.273 & SSIM: 0.582\\
            PSNR: $\infty$ & PSNR: 18.224 & PSNR: 25.88\\   
            
        \includegraphics[width=0.25\textwidth]{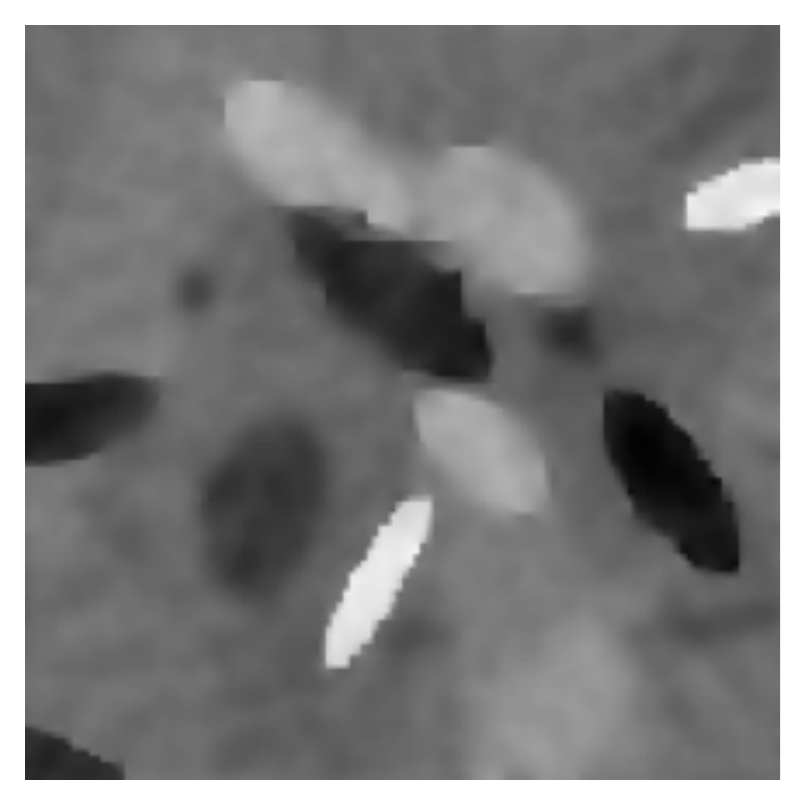}   
        & 
        
        \includegraphics[width=0.25\textwidth]{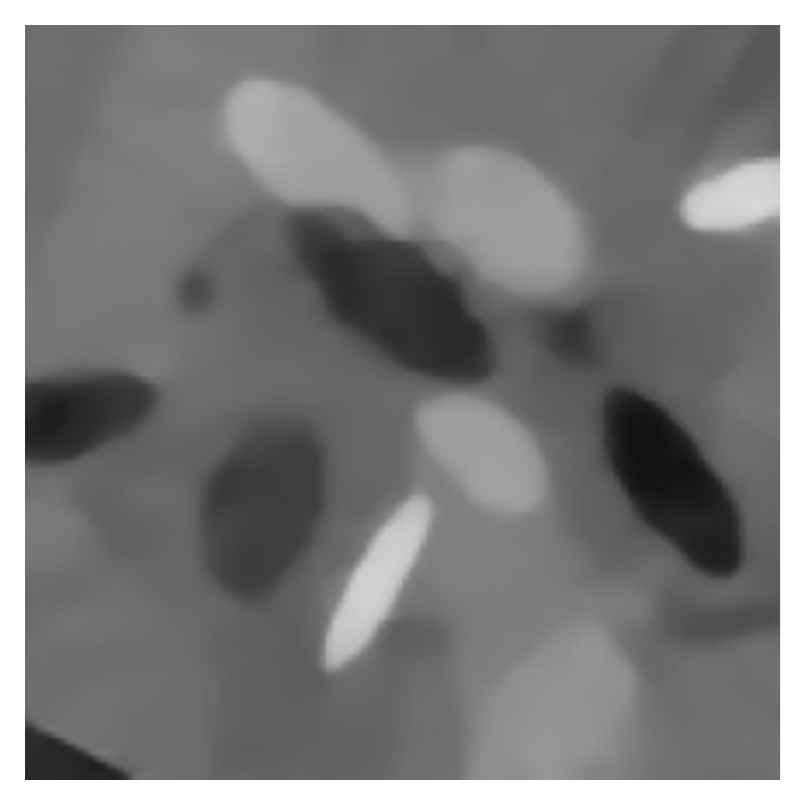}    
        &  
        \includegraphics[width=0.25\textwidth]{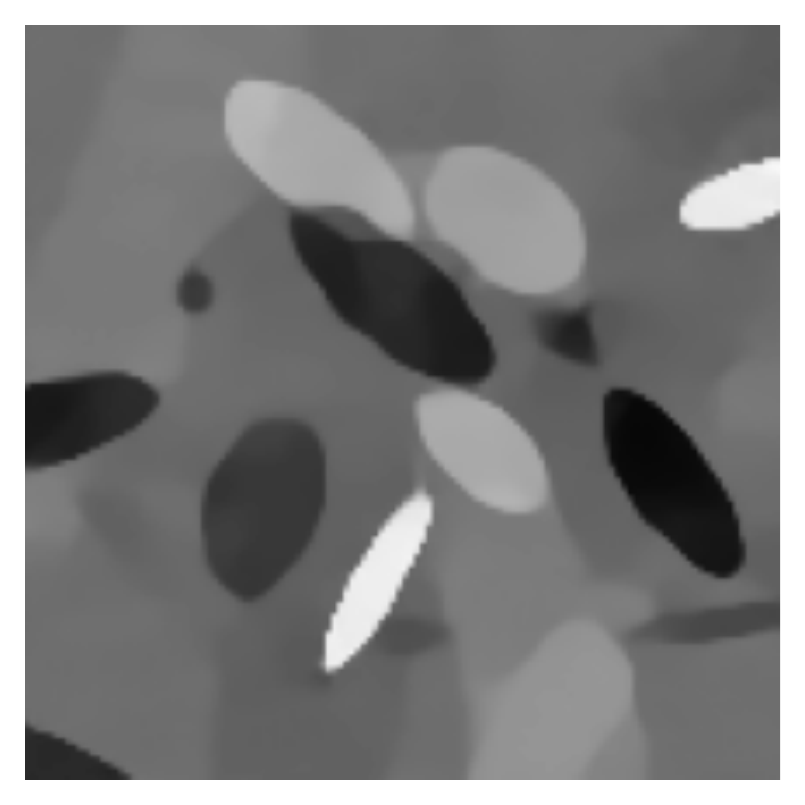}   \\
        
            {\bf TVM} & {\bf Unrolling} &{\bf  F-FPN}
            \\      
            SSIM: 0.786 & SSIM: 0.811 & SSIM: 0.900\\
            PSNR: 27.80 & PSNR: 26.01 & PSNR: 30.94\\
            
        \end{tabular}
        \caption{Ellipse reconstruction with test data for each method: filtered back projection (FBP) TV superiorization  (TVS), TV minimization (TVM), unrolled network, and the proposed feasibility-based fixed point network (F-FPN).}
        \label{fig: ellipses-wide}
    \end{figure}

 \begin{figure}[t]
      \centering
      \small
        \setlength{\tabcolsep}{7pt}
        \begin{tabular}{c c c}    
        
        \includegraphics[width=0.25\textwidth, clip, trim={60 110 60 10}]{./Learned_Feasibility_Ellipse_GT_ind_0000.pdf}       
        &  
        \includegraphics[width=0.25\textwidth, clip, trim={60 110 60 10}]{./Learned_Feasibility_Ellipse_FBP_ind_0000.pdf}      
        & 
        \includegraphics[width=0.25\textwidth, clip, trim={60 110 60 10}]{./Learned_Feasibility_Ellipse_TVS_ind_0000.pdf}    \\
        
            {\bf Ground Truth} & {\bf FBP} &{\bf  TVS}
            \\      
            SSIM: 1.000 & SSIM: 0.273 & SSIM: 0.582\\
            PSNR: $\infty$ & PSNR: 18.224 & PSNR: 25.88\\   
            
        \includegraphics[width=0.25\textwidth, clip, trim={60 110 60 10}]{./Learned_Feasibility_Ellipse_TVM_ind_0000.pdf}   
        & 
        
        \includegraphics[width=0.25\textwidth, clip, trim={60 110 60 10}]{./Learned_Feasibility_Ellipse_Unrolled_ind_0000.pdf}    
        &  
        \includegraphics[width=0.25\textwidth, clip, trim={60 110 60 10}]{./Learned_Feasibility_Ellipse_FFPN_ind_0000.pdf}   \\
        
            {\bf TVM} & {\bf Unrolled} &{\bf  F-FPN}
            \\      
            SSIM: 0.786 & SSIM: 0.811 & SSIM: 0.900\\
            PSNR: 27.80 & PSNR: 26.01 & PSNR: 30.94\\
            
        \end{tabular}
        \caption{Zoomed-in   ellipse reconstruction with test data of Figure \ref{fig: ellipses-wide} for each method:  FBP, TVS, TVM, unrolling,  and the proposed  F-FPN.}
        \label{fig: ellipses-close}
    \end{figure}

\begin{figure}[t]
      \centering
      \small
        \setlength{\tabcolsep}{7pt}
        \begin{tabular}{c c c}     
        
        \includegraphics[angle=180, origin=c,width=0.25\textwidth]{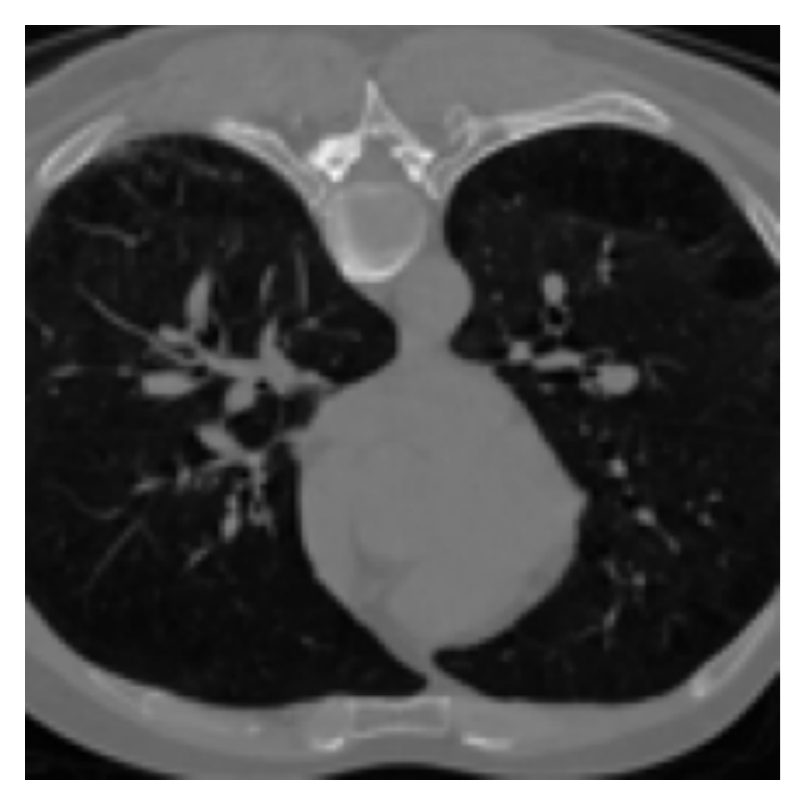}  
         
        & 
        \includegraphics[angle=180, origin=c,width=0.25\textwidth]{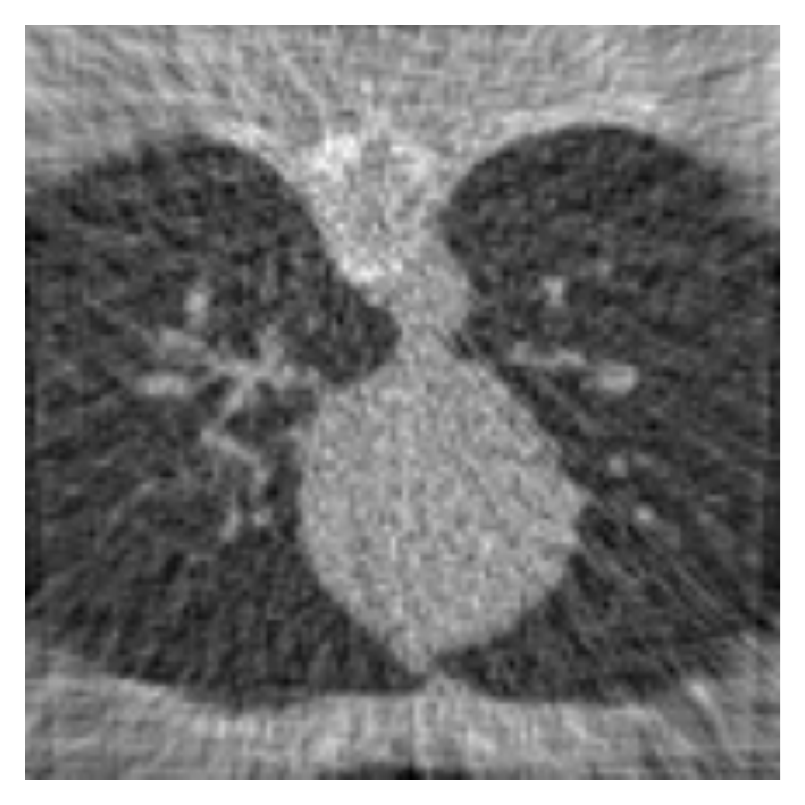}  
         
        & 
        \includegraphics[angle=180, origin=c,width=0.25\textwidth]{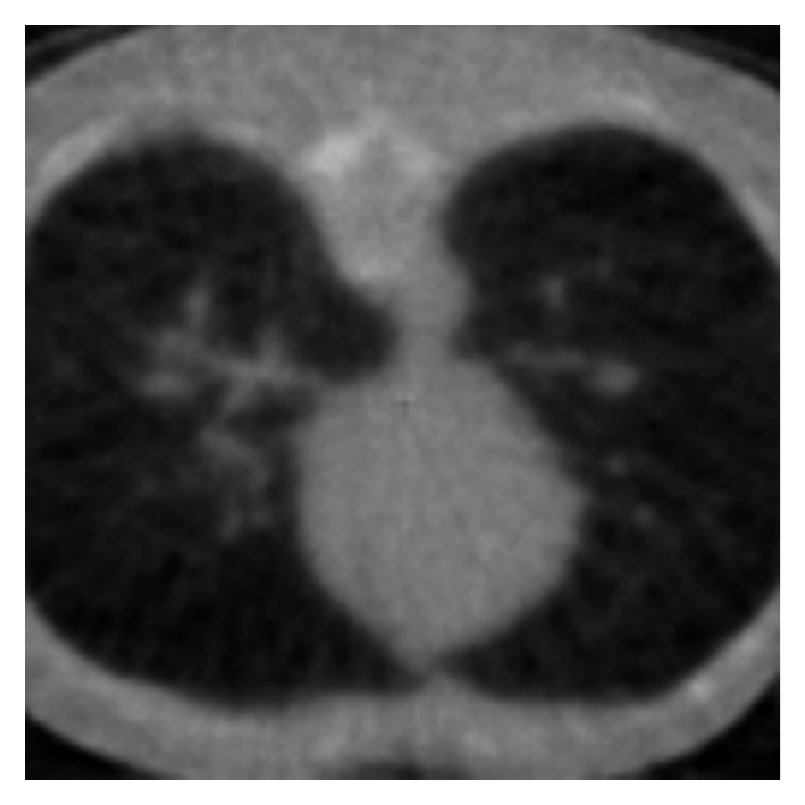}   \\
        
            {\bf Ground Truth} & {\bf FBP} &{\bf  TVS}
            \\      
            SSIM: 1.000 & SSIM: 0.423 & SSIM: 0.686 \\
            PSNR: $\infty$ & PSNR: 18.86 & PSNR: 24.74 \\   
            
            \includegraphics[angle=180, origin=c,width=0.25\textwidth]{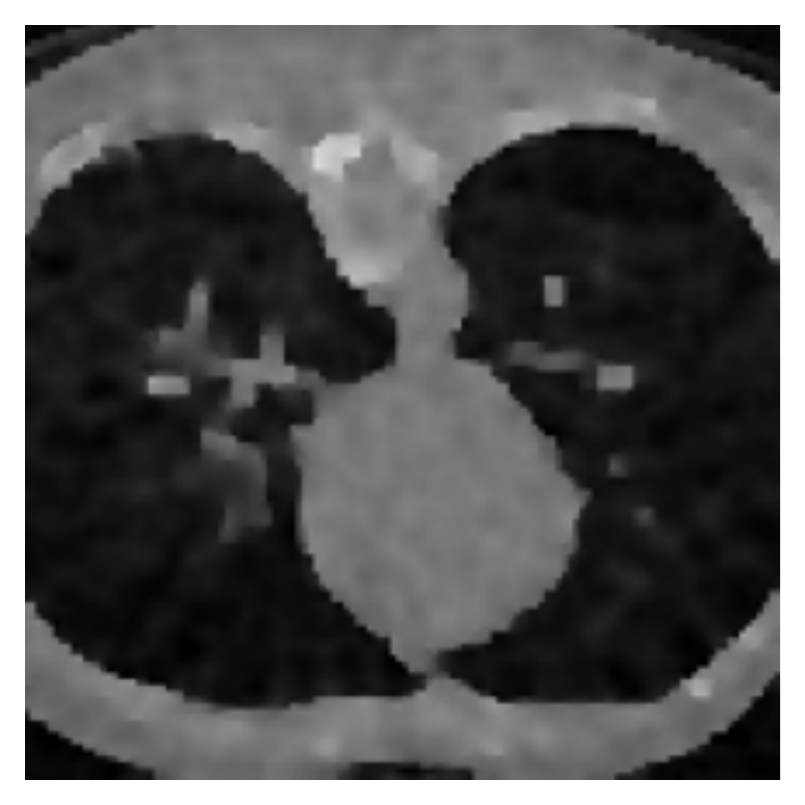}   
        & 
         \includegraphics[angle=180, origin=c,width=0.25\textwidth]{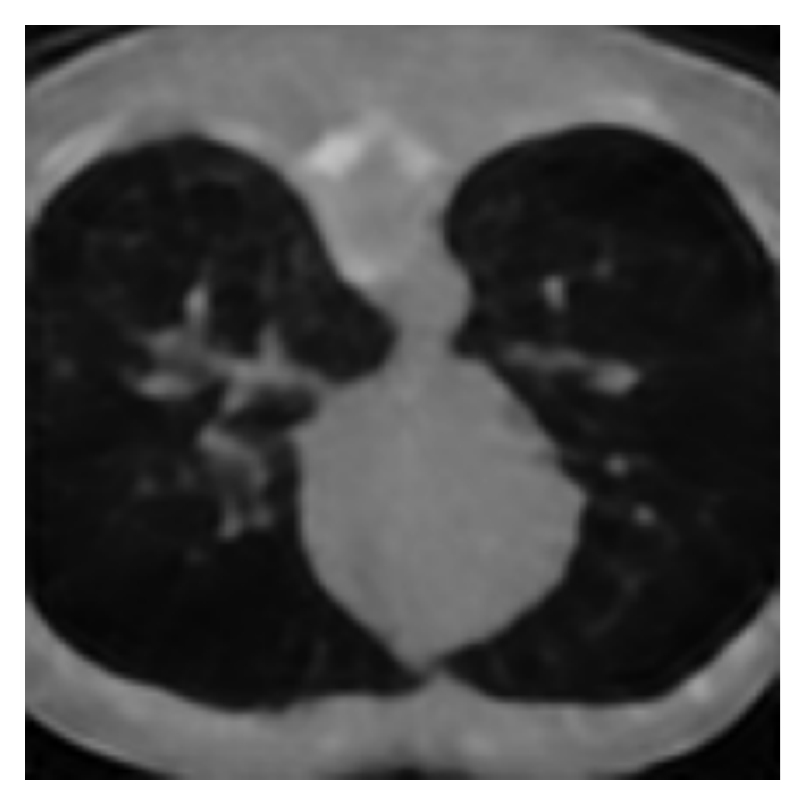}  
         &
          \includegraphics[angle=180, origin=c,width=0.25\textwidth]{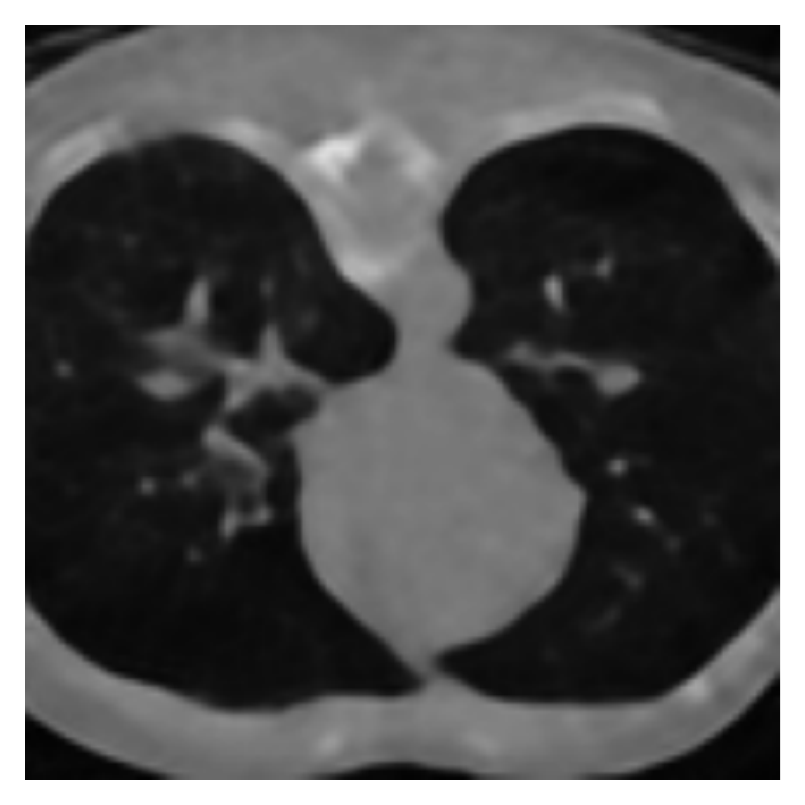}  \\
        
            {\bf TVM} & {\bf Unrolling} &{\bf  F-FPN}
            \\      
            SSIM: 0.761 & SSIM: 0.787 & SSIM: 0.827\\
            PSNR: 26.85 & PSNR: 27.14  & PSNR: 28.82\\
            
        \end{tabular}
        \caption{LoDoPab reconstruction with test data for each method: filtered back projection (FBP) TV superiorization  (TVS), TV minimization (TVM), unrolled network, and the proposed feasibility-based fixed point network (F-FPN).}
        \label{fig: LoDoPab-wide}
    \end{figure}
    
\section{Experiments} \label{sec: experiments}
    Experiments in this section demonstrate the relative reconstruction quality of F-FPNs and comparable schemes -- in particular,  filtered backprojection (FBP)~\cite{dudgeon1990multidimensional},  total variation (TV) minimization (similarly to \cite{oconnor2014primal,goldstein2009split}), total variation superiorization (based on \cite{penfold2010total,humphries2017superiorized}), and an unrolled L2O scheme with an RNN structure. 

 \subsection{Experimental Setup}
    Comparisons are provided for two low-dose CT examples: a synthetic dataset, consisting of images of random ellipses, and the LoDoPab dataset \cite{leuschner2019lodopab}, which consists of human phantoms.
    For both datasets, CT measurements are simulated with a parallel beam geometry with a sparse-angle setup of only $30$ angles and $183$ projection beams, resulting in 5,490 equations and 16,384 unknowns. Additionally, we add $1.5\%$ Gaussian noise corresponding to each individual beam measurement.
    Moreover, the images have a resolution of $128 \times 128$ pixels.  
    The quality of the image reconstructions are determined using the Peak Signal-To-Noise Ratio (PSNR) and structural similarity index measure
    (SSIM).  
    We use the PyTorch deep learning framework \cite{paszke2019pytorch} and the ADAM \cite{kingma2015adam} optimizer. We also use the Operator Discretization Library (ODL) python library \cite{adler2017odl} to compute the filtered backprojection solutions.  
    The CT experiments are run on a Google Colab notebook.
    For all methods, we use a single diagonally relaxed orthogonal projections (DROP) \cite{censor2008diagonally} operator for $\sA_d$  (\ie $\sA_d^k = \sA_d$ for all $k$), noting   DROP is nonexpansive with respect to a norm dependent on $A$ \cite{Heaton2019ASI}. The loss function $\ell$ used for training is the mean squared error between reconstruction estimates and the corresponding true signals.   %\paragraph{Ellipse Phantoms}
    We use a synthetic dataset consisting of random phantoms of combined ellipses as in~\cite{adler2017solving}. The ellipse training and test sets contain 10,000 and 1,000 pairs, respectively. 
%  \paragraph{Human Phantoms}
    We also use phantoms derived from actual human chest CT scans via the benchmark Low-Dose Parallel Beam dataset (LoDoPaB)~\cite{leuschner2019lodopab}.  
    The LoDoPab training and test sets contain 20,000 and 2,000 pairs, respectively.

    \begin{table}[t]
        \centering
        \rmfamily 
        \begin{tabular}{c c c c}
        \toprule
        Method & Avg. PSNR (dB) & Avg. SSIM & \# Parameters 
        \\
        \midrule
        \rowcolor{\ROWCOLOR}
        Filtered Backprojection & 17.79 & 0.211 & 1
        \\ 
        TV Superiorization & 27.35 & 0.721   & 2
        \\
        
        \rowcolor{\ROWCOLOR}
        TV Minimization &  28.55 &  0.772 & 4
        \\        

        Unrolled Network &  30.39 & 0.859 & 96,307
        \\  
        
        \rowcolor{\ROWCOLOR} 
         {\bf F-FPN (proposed)} & {\bf  31.30} &  {\bf 0.877} & 96,307
        \\         
        \vspace{.0mm}
        \end{tabular}
        \caption{Average PSNR and SSIM on the 1,000 image ellipse testing dataset.   }
        \label{tab: ellipse_results}
    \end{table}

    \subsection{Experiment Methods}
    \paragraph{TV Superiorization}  
    Sequences generated by successively applying the   operator $\sA_d$ are known to converge even in the presence of summable perturbations, which can be intentionally added to lower a regularizer   value (\eg TV) without compromising   convergence, thereby giving a ``superior'' feasible point. Compared to minimization methods, superiorization typically only guarantees feasibility, but is often able to do so at reduced computational cost. 
    This scheme, denoted as TVS, generates  updates
    \begin{equation}
        u^{k+1} = \sA_d\left(u^k - \alpha \beta^k  D_{-}^\top \left(\dfrac{D_+u}{\|D_+u\|_2+\varepsilon}\right) \right),
        \ \ \ \mbox{for $k=1,2,\ldots,20$,}
        \label{eq: TVS-update}
    \end{equation}
    where $D_-$ and $D_+$ are the forward and backward differencing operators,   $\varepsilon>0$ is added for stability, and 20 iterations are used as early stopping to avoid overfitting to noise. The differencing operations yield a   derivative of isotropic TV (\eg see \cite{lie2007inverse}). The scalars $\alpha >0 $ and  $\beta \in (0,1)$ are chosen to minimize training mean squared error.
    See the superiorization bibliography \cite{censor2015superiorization} for further TVS materials.
    
    \paragraph{TV Minimization} For a second analytic comparison method, we use anisotropic TV minimization (TVM). In this case, we solve the constrained  problem
    \begin{equation}
        \min_{u\in [0,1]^n} \|D_+u\|_1 \ \ \mbox{such that\ } \ 
        \|Au-d\|\leq\varepsilon,
        \tag{TVM}
        \label{eq: TVM}
    \end{equation}
    where $\varepsilon > 0$ is a hand-chosen scalar reflecting the level of measurement noise  and the box constraints on $u$ are included since all signals have pixel values in the interval $[0,1]$. We use linearized ADMM \cite{ryu2022large} to solve (\ref{eq: TVM}) and refer to  this model as TV minimization (\ref{eq: TVM}). Implementation details are in the Appendix.
    
    \begin{table}[t]
         
        \centering
        \rmfamily   
        \begin{tabular}{c c c c}
        \toprule
        {Method} & Avg. PSNR (dB) & Avg. SSIM & \# Parameters 
        \\
        \midrule
        \rowcolor{\ROWCOLOR}
        Filtered Backprojection & 19.27 & 0.354 & 1
        \\ 
        TV Superiorization &  26.65 & 0.697 & 2
        \\
        
        \rowcolor{\ROWCOLOR}
        TV  Minimization & 28.52 & 0.765 & 4
        \\        

        Unrolled Network &  29.30 & 0.800 & 96,307
        \\  
        
        \rowcolor{\ROWCOLOR} 
        F-FPN (proposed) & {\bf  30.46} & {\bf 0.832} & 96,307
        \\        
        %\bottomrule
        \vspace{.0mm}
        \end{tabular}
        \caption{Average PSNR/SSIM on the 2,000 image LoDoPab testing dataset.}
        \label{tab: LoDoPab_results}
    \end{table} 
 
    \paragraph{F-FPN Structure}
    The architecture of the operator $R_\Theta$ is modeled after the seminal work \cite{he2016deep} on residual networks.
    The F-FPN and unrolled scheme both leverage the same structure $R_\Theta$ and DROP operator for $\sA_d$. 
    The operator $R_\Theta$ is the composition of four residual blocks.
    Each residual block  takes the form of the identity mapping plus the composition of a leaky ReLU activation function and convolution (twice).
    The number of network weights in $R_\Theta$ for each setup   was 96,307,  a small number by machine learning standards.
    Further details are provied  in the Appendix.
    
\begin{figure}[t]
      % trim=<llx> <lly> <urx> <ury>
      \centering
      \small
        \setlength{\tabcolsep}{7pt}
        \begin{tabular}{c c c}     
        \includegraphics[angle=180, origin=c, width=0.25\textwidth, clip, trim={60 110 60 10}]{./Learned_Feasibility_Lodopab_GT_ind_1000.pdf}  
        & 
        \includegraphics[angle=180, origin=c, width=0.25\textwidth, clip, trim={60 110 60 10}]{./Learned_Feasibility_Lodopab_FBP_ind_1000.pdf}  
    
        & 
        \includegraphics[angle=180, origin=c, width=0.25\textwidth, clip, trim={60 110 60 10}]{./Learned_Feasibility_Lodopab_TVS_ind_1000.pdf}   \\
        
            {\bf Ground Truth} & {\bf FBP} &{\bf  TVS}
            \\      
            SSIM: 1.000 & SSIM: 0.423 & SSIM: 0.686\\
            PSNR: $\infty$ & PSNR: 18.86 & PSNR: 24.74\\   
            
            \includegraphics[angle=180, origin=c, width=0.25\textwidth, clip, trim={60 110 60 10}]{./Learned_Feasibility_Lodopab_TVM_ind_1000.pdf}   
        & 
        \includegraphics[angle=180, origin=c, width=0.25\textwidth, clip, trim={60 110 60 10}]{./Learned_Feasibility_Lodopab_Unrolled_ind_1000.pdf}  
         &
          \includegraphics[angle=180, origin=c, width=0.25\textwidth, clip, trim={60 110 60 10}]{./Learned_Feasibility_Lodopab_FFPN_ind_1000.pdf}  \\
        
            {\bf TVM} & {\bf Unrolling} &{\bf  F-FPN}
            \\      
            SSIM: 0.761 & SSIM: 0.787 & SSIM: 0.827\\
            PSNR: 26.85 & PSNR: 27.14 & PSNR: 28.82\\
            
        \end{tabular}
        \caption{Zoomed-in LoDoPab reconstruction with test data of Figure \ref{fig: LoDoPab-wide} for each method:  FBP, TVS, TVM, unrolling,  and the proposed  F-FPN.}
        \label{fig: LoDoPab-close}
    \end{figure}
    
    \subsection{Experiment Results}
    Our results show that F-FPN outperforms all classical methods as well as the unrolled data-driven method.
    We show the result on an individual reconstruction via wide and zoomed-in images from the ellipse and LoDoPab testing datasets in Figures~\ref{fig: ellipses-wide} and \ref{fig: ellipses-close} and Figures \ref{fig: LoDoPab-wide} and \ref{fig: LoDoPab-close}, respectively.
    %We also show the result on an individual figure from the LoDoPab testing dataset in Figure~\ref{fig: LoDoPab-wide} and its corresponding zoomed-in version in Figure~\ref{fig: LoDoPab-close}. 
    The average SSIM and PSNR values on the entire ellipse and LoDoPab datasets are shown in Tables \ref{tab: ellipse_results} and \ref{tab: LoDoPab_results}.  We emphasize  the type of noise depends on each individual ray in a similar manner to~\cite{heaton2020wasserstein}, making the measurements more noisy than some related works. This noise and ill-posedness of our underdetermined setup are illustrated by the poor quality of analytic method reconstructions. (However, we note improvement by using TV over FBP and further improvement by TV minimization over TV superiorization.)
    Although nearly identical in structure to F-FPNs, these results show the unrolled method to be inferior to F-FPNs in these experiments. We hypothesize this is due to the large memory requirements  of unrolling (unlike F-FPNs), which limits the number of unrolled steps ($\sim 20$ steps versus 100+ steps of F-FPNs), and F-FPNs are tuned to optimize a fixed point condition rather than a fixed number of updates.

 \section{Conclusion}\label{sec: conclusion}
This work connects feasibility-seeking algorithms and data-driven algorithms (\ie neural networks). The F-FPN framework leverages the elegance of fixed point methods while using state of the art training methods for implicit-depth deep learning. This results in a sequence of learned operators $\{\sA_d^k\circ R_\Theta\}_{k\in\bbN}$ that can be repeatedly applied until convergence is obtained. This limit point is expected to be nearly compatible with provided constraints (up to the level of noise) and resemble the collection of true signals. The provided numerical examples show improved performance obtained by F-FPNs over both classic methods and an unrolling-based network. Future work will extend FPNs to a wider class of optimization problems and further establish theory connecting machine learning to fixed point methods. 
 
\newpage

\section{Appendix}
  
\subsection{Network Structure}
    For our neural network architecture, we set $R_\Theta$ to be a composition of four convolutions: the first takes in one channel and outputs 44 channels. The   second and third convolutions have 44 input and output channels. The final convolution maps the 44 channels back to one channel. Prior to each convolution, we use the leaky rectified linear activation function (ReLU) as the nonlinear activation function between layers. The leaky ReLU function, denoted by $\phi$, is defined as
    \begin{equation}
        \phi_a (u) \triangleq \begin{cases}
        \begin{array}{cl}
            u & \mbox{if $u \geq 0$,} \\
            a u & \mbox{if $u < 0$.} 
        \end{array}
        \end{cases}
    \end{equation}
    where $a$ is a number determined by the user. Exact implementation details can be found in \url{https://github.com/howardheaton/feasibility_fixed_point_networks}.

\subsection{Training Setup}
To generate the FBP reconstructions, we use FBP operator from the Operator Discretization Library (ODL). Since the ODL FBP operator is a built-in operator whose rows are not normalized (unlike in the remainder of the methods where DROP is used), we scale the observed data accordingly. In particular, we multiply each row of the observed data $d$ by the rows of the original, unnormalized matrix $A$. For all other methods, we normalized the rows of $A$ and scaled the measurements accordingly. \\

For the ellipse dataset, we train the unrolled network using a batch size of 15 for 60 epochs. The F-FPN network training used a batch-size of 15 for 50 epochs. The unrolled network architecture contains 20 total layers (\ie update steps) -- the number of layers was chosen on the memory capacity of the GPU.\\

For the LoDoPab dataset, we train the unrolled and F-FPN networks using a batch-size of 50 for 50 epochs total. The unrolled network architecture contains 14 total layers (\ie update steps) -- the number of layers was chosen on the memory capacity of the GPU.\\

\subsection{TVS parameters}
The TVS parameters were trained by unrolling the method indicated in (\ref{eq: TVS-update}) for 20 steps into the structure of a neural network. This unrolled network contained 2 parameters: $\alpha$ and $\beta$. We initialized $\alpha$ to $0.05$ and $\beta$ to $0.99$. Then we used Adam to tune the parameters with the training data. For the ellipse experiment, the learned parameters were $\alpha = 0.0023$ and $\beta = 0.968$. For the LoDoPab experiment, the learned parameters were $\alpha = 0.0013$ and $\beta = 9607$. Note the training to tune the parameters optimized performance with respect to mean squared error.

\subsection{Approximate Lipschitz Enforcement} \label{subsec: approximate-contraction}
Herein we overview our technique for ensuring the composition $(\sA_d\circ R_\Theta)$ is $\gamma$-Lipschitz in our experiments (with $\gamma \in (0,1]$). This is accomplished in an approximate manner using batches of computed fixed points after each forward pass in training. Let $B$ denote a set of indices corresponding to a  collection of fixed points $\tilde{u}_d$ and let $\{\zeta_i\}_{i\in B}$ be Gaussian random vectors. Letting $|B|$ denote the cardinality of $B$, we check whether the following equality holds:
\begin{equation}
    \underbrace{\dfrac{1}{|B|}\sum_{i\in B} \left\| (\sA_d\circ R_\Theta)(\tilde{u}_d) - (\sA_d\circ R_\Theta)(\tilde{u}_d + \zeta_i ) \right\| }_{C_1}
    \leq \gamma \underbrace{\dfrac{1}{|B|} \sum_{i} \|\zeta_i\|}_{C_2}.
    \label{eq: contraction-batch}
\end{equation}
If the network is $\gamma$-Lipschitz, then $C_1 \leq \gamma C_2$ for any provided batch $B$ of samples. 
Now suppose the inequality does not hold, the case where action must be taken. First assume $\sA_d$ is $1$-Lipschitz. Then it suffices to make $R_\Theta$ $\gamma$-Lipschitz. As noted previously, $R_\Theta$ takes the form of a composition of ResNet blocks. For simplicity, suppose
\begin{equation}
    R_\Theta = \mathrm{I} + \phi_a (Wu + b),
\end{equation}
for a matrix $W$ and vector $b$ defined in terms of the weights $\Theta$. 
Let $C_3 \triangleq \gamma C_1 / C_2$.
To make (\ref{eq: contraction-batch}) hold, it would be sufficient to replace $R_\Theta$ by $ C_3 \cdot R_\Theta$. 
Furthermore,
\begin{subequations} 
\begin{align}
    C_3 R_\Theta 
    &= C_3  \left( \mathrm{I} + \phi_a(Wu+b)\right)\\
    &= \mathrm{I} + C_3 \phi_a(Wu+b) + \left( C_3 - 1\right)\mathrm{I}\\
    & \approx \mathrm{I} +  \phi_a\left( C_3 (Wu+b)\right)+ \left( C_3 - 1\right)\mathrm{I}\\   
    & \approx \mathrm{I} +  \phi_a\left( C_3 (Wu+b)\right),  
\end{align}
\end{subequations}
where the first approximation is an equality when $Wu+b \geq 0$ (and approximately equal when $a$ is small), and the second approximation holds whenever the inequality (\ref{eq: contraction-batch}) is ``close'' to hold, \ie  $C_3 \approx 1$. This shows that
\begin{equation}
    C_3 R_\Theta \approx  \mathrm{I} +  \phi_a\left( C_3 (Wu+b)\right).
\end{equation}
Thus, to ensure $R_\Theta$ is approximately $\gamma$-Lipschitz, we may do the following. After each forward pass in training (\ie computing $\sN_\Theta(d)$ for a batch $B$ of data $d$), we compute $C_1$ and $C_2$ as above. If (\ref{eq: contraction-batch}) holds, then no action is taken. If (\ref{eq: contraction-batch}) does not hold, then multiply the weights $W$ and $b$ by $C_3$, making (\ref{eq: contraction-batch}) hold. \\
 
In our experiments, the structure of $R_\Theta$ was a more complicated variation of the above case (namely, the residual portion was the composition of convolutions). However, we used the same normalization factor, which forces $R_\Theta$ to be slightly more contractive than needed.
And, in the general case where $R_\Theta$ is the composition of mappings of the form identity plus residual, it suffices to multiply the weights by the normalization constant $C_3$ raised to one over the number of layers $\ell$ in the residual mapping (\ie $C_3^{1/\ell}$). 

\begin{remark}
    An important note must be made with respect to normalization. Namely, $R_\Theta$ was almost never updated by the procedure above. Because of the initialization of the weights $\Theta$, $R_\Theta$ appears to have been roughly 1-Lipschitz. And, because the weights are tuned to improve the performance of $R_\Theta$, it appears that this typically resulted in updates that did not make $R_\Theta$ less contractive. Consequently, the above is an approximate safeguard, but did not appear necessary in practice to obtain our results.
\end{remark}

\subsection{TV Minimization}

We equivalently rewrite the problem (\ref{eq: TVM}) as
\begin{equation}
    \min_{u,p,w} \delta_{[0,1]^n}(u) + \|p\|_1 + \delta_{B(d,\varepsilon)}(w)
    \ \ \ \mbox{such that}\ \ \ 
    \left[\begin{array}{c} D_+ \\ A\end{array}\right] u - \left[\begin{array}{c} p \\ w \end{array}\right] = 0,
    \label{eq: TVM-substitution-1}
\end{equation}
where $D_+$ is the concatenation of forward difference operators along each image axis.
Using a change of variables $\xi = (p,w)$, defining the function
\begin{equation}
    f(\xi) \triangleq \|p\|_1 + \delta_{B(d,\varepsilon)}(w),
\end{equation}
and setting $M = [D_+; A]$, we rewrite (\ref{eq: TVM-substitution-1}) as
\begin{equation}
    \min_{u,\xi} \delta_{[0,1]^n}(u) + f(\xi)
    \ \ \ \mbox{such that}\ \ \ 
    Mu-\xi = 0.
    \label{eq: TVM-substitution-2}
\end{equation}
Observe (\ref{eq: TVM-substitution-2}) follows the standard form of ADMM-type problems. For scalars $\alpha,\beta,\lambda \in(0,\infty)$, linearized ADMM \cite{ryu2022large} updates take the form
\begin{subequations}
\begin{align}
    u^{k+1} & = P_{[0,1]^n}\left( u^k - \beta M^\top(\nu^k + \alpha (Mu^k-\xi^k))\right), \\
    \zeta^{k+1} & = \prox{\lambda f}\left(\xi^k + \lambda(\nu^k + \alpha (Mu^{k+1}-\xi^k))\right), \\
    \nu^{k+1} & = \nu^k + \alpha (Mu^{k+1} - \xi^{k+1}).
\end{align}
\end{subequations}
Expanding terms, we obtain the explicit formulae
\begin{subequations}
\begin{align}
    r^{k} & = D_+^\top (\nu_1^k + \alpha (D_+u^k - p^k)) + A^\top (\nu_2^k + \alpha (Au^k-w^k)),\\
    u^{k+1} & = P_{[0,1]^n}\left( u^k - \beta  r^k\right) ,\\
    p^{k+1} & = \eta_\lambda \left(p^k + \lambda(\nu_1^k + \alpha (D_+u^{k+1}-p^k))\right), \\
    w^{k+1} & = P_{B(d,\varepsilon)} \left(w^k + \lambda(\nu_2^k + \alpha (Au^{k+1}-w^k))\right), \\    
    \nu_1^{k+1} & = \nu_1^k + \alpha (D_+u^{k+1} - p^{k+1}),\\
    \nu_2^{k+1} & = \nu_2^k + \alpha (Au^{k+1} - w^{k+1}),
\end{align}
\end{subequations}
where $B(d,\varepsilon)$ is the Euclidean ball of radius $\varepsilon$ centered at $d$ and $\eta_\lambda$ is the soft thresholding operator with parameter $\lambda$, \ie
\begin{equation}
    \eta_\lambda(u) \triangleq \begin{cases}
    u - \lambda & \mbox{if $x\geq \lambda$,} \\
    u + \lambda & \mbox{otherwise.}
    \end{cases}
\end{equation}
We set $u^1 = 0$, $\nu^1 = 0$, $p^1 = D_+ u^1$, and $w^1 = Au^1$.
For the ellipses experiment, we use $\alpha=\beta=\lambda=0.1$, $\varepsilon = 10$, and 250 iterations. 
For the LoDoPab experiment, we use $\alpha=\beta=\lambda=0.1$, $\varepsilon = 5$, and 250 iterations.
Note the computational costs of computing each signal estimate via TVM is  greater than FBP and TVS.

% \subsection{Additional Images}

%%%%%%%%%%%%%%%%%%%%%%%%%%%%%%%%%%%%%%%%%%%%%%
%%                                          %%
%% Backmatter begins here                   %%
%%                                          %%
%%%%%%%%%%%%%%%%%%%%%%%%%%%%%%%%%%%%%%%%%%%%%%

\begin{backmatter}

\section*{Acknowledgements}%% if any
We thank Daniel Mckenzie and Qiuwei Li for their helpful feedback prior to submitting our paper.

\section*{Funding}%% if any
Samy Wu Fung is supported by AFOSR MURI FA9550-18-1-0502, AFOSR Grant No. FA9550-18-1-0167, and ONR Grants N00014-18-1-2527 snf N00014-17-1-21. 
Howard Heaton is supported by the National Science Foundation (NSF) Graduate Research Fellowship under Grant No. DGE-1650604. Any opinion, findings, and conclusions or recommendations expressed in this material are those of the authors and do not necessarily reflect the views of the NSF.

% \section*{Abbreviations}%% if any
% Text for this section\ldots

\section*{Availability of data and materials}%% if any
All data can be downloaded in the following link: \href{https://drive.google.com/drive/folders/1Z0A3c-D4dnrhlXM8cpgC1b7Ltyu0wpgQ?usp=sharing}{https://drive.google.com/drive/folders/1Z0A3c-D4dnrhlXM8cpgC1b7Ltyu0wpgQ?usp=sharing}.
The data can also be accessed in the github link where the code is provided.

% \section*{Ethics approval and consent to participate}%% if any
% Text for this section\ldots

\section*{Competing interests}
The authors declare that they have no competing interests.

% \section*{Consent for publication}%% if any
% All authors read and approved the manuscript.

\section*{Authors' contributions}
All authors contributed equally and significantly in this research work. All authors read and approved the final manuscript.

% \section*{Authors' information}%% if any
% Text for this section\ldots

%%  After submission of the .TEX file,                     %%
%%  you will be prompted to submit your .BBL file.         %%

% if your bibliography is in bibtex format, use those commands:
\bibliographystyle{bmc-mathphys} % Style BST file (bmc-mathphys, vancouver, spbasic). 
\bibliography{ffpn_references}      % Bibliography file (usually '*.bib' )

\end{backmatter}
\end{document}